# Grammatical vs Spelling error correction: An investigation into the responsiveness of Transformer based language models using BART and MarianMT

Rohit Raju[1,2], Peeta Basa Pati[*,2], SA Gandheesh[2], Gayatri Sanjana Sannala[2] & Suriya KS[2]

[1]University of Colorado Boulder, CO, US, e-mail: rohit.raju@colorado.edu

[2]Department of Computer Science & Engineering, Amrita School of Computing Bengaluru, Amrita Vishwa Vidyapeetham, INDIA

*Corresponding Author: bp_peeta@blr.amrita.edu, ORC ID: *0000-0003-2376-4591*

**Abstract**

Text continues to remain a relevant form of representation for information. Text documents are created either in digital native platforms or through conversion of other media files such as images and speech. While the digital native text is invariably obtained through physical or virtual keyboards, technologies such as OCR & speech recognition are utilized to transform the images and speech signals to text content. All these variety of mechanisms of text generation also introduce error into the captured text.

This project aims at analyzing different kinds of errors that occurs in text documents. The work employs two of the advanced deep neural network based language models, namely, BART and MarianMT, for rectifying the anomalies present in text. Transfer learning of these models with available dataset is performed to finetune their capacity for error correction. A comparative study is conducted to investigate the effectiveness of these models in handling each of the defined error categories. It is observed that while both the models are able to bring down the erroneous sentences by 20+%, BART is able to handle spelling errors far better (24.6%) than grammatical errors (8.8%).

## I. Introduction

Text is a natural representation of all the existing languages in the world. Texts help one express and communicate with others. Handwritten texts have been part of the history for ages, while digital texts have evolved to keep up with the rapidly growing technology in day to day lives. It is due to texts that one can extend from their knowledge and memory beyond their body into the environment around [1]. Text is available in various forms, from handwritten manuscripts to digitally written blogs, from stone carvings to printed posters. Texts can be utilized for personal reasons such as diary entry, blog, etc., as well as for professional purposes like advertising, surveying, etc. Right from the newspaper one reads in the morning to the social media scrolling before going to bed, people are surrounded by text.

It is human nature to categorize any kind of data they receive. As there is so much text available around, it is obvious that humans tend to inspect and review the text they require. Thus, the origin of text analysis. It is the process of scanning the textual data in order to derive some meaning and store information.

Most businesses rely on text analysis to extract valuable insights from various raw sources. The feedback received from these sources such as emails, chat messages, social media posts, comments & statements and survey responses help them in their decision-making strategies. Text capturing can be complex and tend to introduce new errors based on its source and capture technology.

Besides digital native documents, text may be obtained from other forms of media such as images, video, speech or voice. OCR extracts text present in images, thereby enabling editing and reviewing of the content [2]. The captured textual information may contain errors due to a variety of reasons. Some of these errors are pre-existing in the input images due to incorrect typing or even inaccurate language knowledge. Moreover, the models used to recognize the text in images may introduce some errors, since all OCR engines are inherently error-prone. These recognition errors may occur due to noisy and unclear images, poor handwriting, formatting and spacing issues. Speech recognition [3] is a speech-to-text technology that recognizes spoken words and into its text equivalent. Auto-transcription of speech may introduce errors while phoneme to grapheme conversion is carried out. These errors may be attributed to factors such as substitution of a word with a different, yet similar sounding word, deletion of incoherent words and insertion of contextually irrelevant words. Technologies such as web scraping [4], email-readers and file handlers may also introduce errors while capturing text due to format incompatibility.

Thus, there are mainly two distinct error types available in text sentences, namely, spelling or typographic error and grammatical error. Both these kinds of error may be attributed to errors in symbol recognition, mis-typing and incorrect language knowledge. Existence of these errors may lead to undesirable consequences, especially in fields such as courtrooms & hospitals. For example, the sentences "The accused fled from the crime" and "The accused bled from the crime" can lead to very different outcomes for the case based on just one spelling mistake. The sentence "He are healthy" is grammatically incorrect, yet highly possible to be created by someone who lacks proper knowledge of English grammar.

There are several methods to eradicate these errors. While one can utilize manual labor to individually pick out the errors, it can be time consuming and can add some human errors. Rule based error correction [5] is also a solution to eliminate error. In such a system, first a large number of documents are studied to identify the common error patterns based one which the rules are defined. These defined rules are, subsequently, employed to bring corrections to errors in future documents. Efficiency of such error correction is based on the quality of defined rules as well as ordering and properties of the rules. Though it is useful in reducing the errors in text, it is sensitive to newer language structures. Rule Based error correction is often time consuming, without any learning capacity and is very unidimensional, thereby, making it inefficient. To overcome these shortcomings, advanced deep neural network based Natural Language Processing (NLP) models are used for error correction. These NLP models use Artificial Intelligence (AI) and enable machines to read, understand & analyze the meaning as well as context of the text sentences. Since text is a sequential data consisting of a sequence of words, Recurrent Neural Network (RNN) architectures are utilized to analyze and capture the information storing them as language models. Long Short-Term Memory (LSTM) networks are advanced RNN architectures which allow information to persist for relatively longer duration in the model's network. The encoder-decoder architecture of the LSTM carries the capability to learn the context of a sentence and store information in the model, thereby making it a perfect tool for error correction [6]. This paper speaks about two such deep learning NLP models, Bidirectional Auto-Regressive Transformers (BART) [7] and MarianMT [39].

BART and MarianMT models have an encoder-decoder architecture, they work using sequence-to-sequence modeling of language data, and are memory-based. During training, these models pick up on the traits and rules of the grammar. The noisy text is fed as input to these language models, which produce accurate sentences as output. This helps to improve the accuracy of generated text. Both these models accomplish the stated task by comprehending the grammar and context of used words in a sentence. Token masking, token detection, text infill & sentence permutation are tasks that the BART and MarianMT decoders are trained to perform.

BART is an advanced language model and has been reported for its efficacy for various language processing tasks. MarianMT is yet to be thoroughly explored in these aspects. For the purpose of this work, both these models are employed to correct various errors present in the input sentence. Many researchers have already employed such language models for accuracy enhancements and reported success with the approach. However, most of these works deal with study of the aggregated quantitative impact, namely, accuracy improvement, error reduction etc. It's equally important to find the pattern of improvement brought in by these models. The outcome of such a study shall help in identification of error scenarios and find the right model to be employed.

In this work, an error category definition is developed & utilized to categorize all the input sentences in the dataset as well as the corresponding predicted outputs. While it's ideal to assume that all input sentences shall be predicted as fully corrected, in practice predicted sentences may also have errors. Thus, it's important to study the shift pattern of the error categories which is the main focus of this work. Observing and analyzing the shift in error categories provides insights into the model's behavior as well as capacity for capturing the context and meaning of the sentences. Such analysis also helps to develop effective strategies to engage such models for appropriate correction based on the types of error available in input sentences. This work aligns with United Nations' Sustainable Development Goals of "Quality Education (SDG-4)" and "Decent Work and Economic Growth (SDG-8)".

The Novelties of this work are as follows:

- Error Category Definition.
- Employ different NLP models and evaluate their capability for category specific error correction and quantity of error present.
- Analysis of percentage of error category shift to understand the capabilities and limitations of the model.
- New subset of data created using manual correction.

This paper is organized as follows: the literature survey is provided in Section II, while Section III provides a definition of various error categories (one of our novelties of this work). The Dataset is discussed in detail in Section IV, Model Design is provided in Section V, analysis of obtained results in Section VI, followed by Conclusion & future scope of the work in Section VII.

## II. Literature Survey

Language models have been employed in numerous research studies on a variety of documents to get a wide range of results. Following are some of the noteworthy research contributions.

A system to improve handwritten paper recognition was proposed by Kumar and Pati [8] using pre-trained models like MarianMT, BERT, and BART. The accuracy of the NLP models was measured using the C4 (trained on 3,000 sentences), Block, Word, and AVV 40 datasets. The authors performed a comparative examination of progress on various language models for various datasets and noted that BART provided the most uplift.

The authors Alikaniotis and Raheja [9] propose that state-of-the-art language models can be used to achieve competitive performance on the task of grammatical error correction (GEC) without the need for annotated training data. The authors reported using pre-trained Transformer language models like BERT, GPT, and GPT-2 to perform GEC & demonstrated that these models create tough baselines to beat. Their work proved that transformer based language-models are effective and robust for the task of GEC, even in absence of annotated training data.

A detailed survey work by Zhang et al. [10] reports about the sensitivity of language models built on deep neural networks towards Adversarial Attacks (the text is bombarded with imperceptible words). Pruithi et al. [11] employed BERT & RNN models to handle Adversarial

Spelling Mistakes and reported accuracies of 90.3% & 75%, respectively. However, a combined BERT and RNN model proved to be quite ineffective for Adversarial attacks and the accuracy drastically dropped to 45.8%.

ERNIE is a model proposed by Zhang et al. [12], which expands the existing abilities of BERT by training the model with knowledge graphs. These knowledge graphs capture rich structured knowledge facts for better language understanding. The model's novelty lies in accurate training with smaller datasets. Kantor et al. [13] proposed a system that combines multiple GEC systems using a black box approach. This work detects the strength of a system or a combination of several systems per error type. The proposed approach has proven that such a system outperforms the average accuracy of ensemble of existing RNN models.

An open-source platform called NeuSpell was created by Jayanthi et al. [14] to simplify correction of spelling mistakes. A wide variety of models, including BERT, SC-LSTM, CHAR-CNN-LSTM, and CHAR-LSTM-LSTM, are available on this platform. Hangaragi et al. [16] utilized the BERT, SC-LSTM, CHAR-CNN-LSTM and CHAR-LSTM-LST models from the NeuSpell to achieve error correction for recognized output from handwritten document images. In their work, Google Vision API (OCR) is used on handwritten documents from the IAM dataset [17], to generate the text data. They reported that BERT provides the best improvement (9.2% at character level) to erroneous text when compared to other models.

Errors introduced by non-native English speakers are a big challenge in Grammatical error correction as there is change in the meaning of the sentences. Liang et al.'s [15] proposed a solution to overcome this problem. They first defined the categories of noise in an English sentence and then injected targeted noise into sentences to build training sets. Subsequently, they fine-tuned BERT with the training set which outperformed various state-of-the-art language models such as LSTM and CNN.

Saluja et al. [18] developed an LSTM-based model with a fixed latency that can learn, detect, and correct OCR errors. Three studies were carried out: error detection (using various evaluation), error correction, and suggestion generation. The results showed that their LTSM model provided the best error correction for Hindi and Malayalam compared to the prior reported models. Introduction of the fixed delay to LSTM is a novelty, different from the standard LSTM models, that enables learning from the subsequent sequence of characters.

Tan et al.'s [19] study focused mostly on BERT for Mandarin spelling error correction. The BERT model, which was trained on a Chinese dataset, is used in the paper to fix incorrect strings. The detection network and the correction network were kept as two distinct parts of the model during design. The task of the detection network is to identify incorrect text (such as misspelled words) while the correction network rectifies the incorrect text. The model's precision, recall, and F1 scores were recorded and compared with those of Kenlm, RNNLM, and BERT-Fintune. An average improvement of 13% for each of these evaluation metrics was observed.

BERT, additionally, has been employed for many other text processing tasks such as summarization of braille documents [41], dependency parsing for Tamil langauge [42] & aspect term extraction for sentiment analysis [43.

Chinese Spelling Check (CSC) is a challenging task due to the complex characteristics of Chinese characters. The most common errors present in Chinese language are phonological or visual errors. To overcome these problems, Huang et al. [20] proposed a novel end-to-end trainable model called PHMOSpell, which promotes the performance of CSC with multi-modal information. Pinyin and glyph graphical representations are derived which are then integrated into a pre-trained language model by a well-designed adaptive gating mechanism. The reported model significantly outperformed all previous state-of-the-art models on precision, recall and f1 score metrics.

Existing state-of-the-art methods either only use a pre-trained language model or incorporate phonological information as external knowledge. To overcome this drawback, Zhang et al. [21] proposed an end-to-end Chinese spelling correction (CSC) model that integrates phonetic features. Initially, the words were replaced with phonetic features and their sound-alike words. Then these words were jointly trained for error correction and detection. The transformer model was trained on SIGHANI15 dataset and significantly outperformed previous state-of-the-art methods with a precision of 77.5%, recall of 83.1% and f1-score of 80.2%.

Xu1 et al. [22] proposed a Chinese spell checker model called ReaLiSe by directly leveraging the multimodal information of the Chinese characters. The ReaLiSe model tackles the CSC task by first capturing the semantic, phonetic, & graphic information of the input characters and then selectively mixing the information in these modalities to predict the correct output. The performance of ReaLiSe model was reported with an accuracy of 84.7%, precision of 77.3%, recall of 81.3% and f1-score of 79.3%.

Kai Fu et al. [23] in this paper detail the approach taken by them to build a Chinese Grammatical Error Correction system. Essays written by the non-native Chinese speakers were used as data and the error correction was carried out in various stages. At the first stage they employed a spelling error correction model which removed the spelling errors. This also acts as a pre-processing step which reduces perturbation at later stages. In the second stage, they cast the grammatical error correction problem as a machine translation task. Here, a sequence-to-sequence model is employed. To achieve this, they experimented with several models with different configurations.

Existing GEC systems suffer from not having enough labeled training data to achieve high accuracy. To overcome this problem, Zhang et al. [24] proposed a copy-augmented architecture for the GEC task by copying the unchanged words from the source sentence to the target sentence. The copy-augmented architecture was pre-trained with unlabeled One Billion Benchmark dataset. This is followed by comparisons between the fully transferred learnt model and a pretrained model. A copying mechanism was applied on the GEC system, which enables the model to copy tokens from the source sentence. The model was evaluated against CoNLL-2014 and JFLEG datasets. The copy-augmented model reported an aggregated precision of 68.48%, recall of 33.10%, f1-score of 56.42% and GLEU score of 59.48 with respect to CoNLL-2014 and GLEU score of 59.48 with respect to JFLEG. The authors reported an increase in the evaluation metrics when the model is combined with denoising auto-encoders. A precision of 71.57%, recall of 38.65%, f1-score of 61.15% with respect to CoNLL-2014 and GLEU score of 61.00 with respect to JFLEG was achieved with such a combination.

Mounika et al. [25] have experimented with both pre-trained and fine-tuned T5 models for lowering the word error rate (WER) [26] in text generated by recognition of speech samples of mathematical equations. On both the models, GloVe & FastText embeddings were used to increase the accuracy. The WER was successfully lowered from 36% to 16% by the proposed system.

Bryant and Briscoe [27] discuss the use of language models (LM) in GEC. The work proposes a simple 5-step approach that relies on very little annotated data and can be used for any language. The approach involves calculating the probability of input sentences, building a confusion set for each token, rescoring the sentence, applying the best correction, and iterating. The work is concluded by highlighting the potential of LM-based approaches for GEC and their usefulness as a baseline for future research.

Sreevidhya & Narayanan [28] have employed three vectorization techniques, namely, Latent Semantic Indexing (LSI), Sentence-BERT & Word2Vec to evaluate the answers by students. These answers are compared against the model answer using these embedding techniques. They employ Cosine similarity to measure the resemblance of the student answer to the model answer.

Based on the survey of the related works, it is inferred that models such as BERT & BART are used not only for various kinds of text processing works but has also been studied for their effectiveness in correction of errors in text. MarianMT is also an enhanced version of BERT and is demonstrated to be effective for spelling error correction. Rohit et al. [40] developed a system to correct anomalies present in English statements using MarianMT and BART. The reported a WER reduction of 34% with MarianMT. Finally, they suggested a systematic analysis to identify the patterns of corrections being performed by each of these models. Therefore, BART and MarianMT are chosen for this work. Additionally, while most reported works have studied the effectiveness of various models for error correction of one type of error (either grammatical or spelling), this work tries to simultaneously handle both these error types. This work, an extension of the work by Rohit et al. [40], spotlights the error shifts occurring with the model predictions to understands the models' capabilities under differing scenarios of anomalies. This helps identify the models' strengths and weaknesses.

### III. Error category definition & shift analysis

Hossain et al. [29] studied the different errors in textual documents and presented the list of eight error types. These error types are: Typographic error, Cognitive error, Visual error, Run-on error, Split-word error, Non-word error, Real-word error. Subsequently, Mounika et al. [25] studied the error types present in automatically speech recognized (ASR) text in their work. Their work provided a list of errors observed in sentences. Both the works of Hossain et al. & Mounika et al. dealt with errors at word level. In this study, these categories of errors are abstracted to sentence level and four distinct categories of errors are proposed. The detailed definitions of these four proposed error categories, namely, Cat A, Cat B, Cat C and Cat D, are provided below. These error categories, each with a short description and example sentences, are tabulated in Table 1.

1. **Cat A:** This is a no-error category. Here, the input sentence matches perfectly with the target sentence.

2. **Cat B:** This category deals with sentences containing errors such as grammatical, word omission, capitalization. The input sentences do not match the target perfectly but all constituent words of the input sentence are valid words. Here, the erroneous words are of real-word error types as described by Hossain et al. Additionally, new words may be added or some words may be missing in the sentence. The various caused which leads to formation of this error category are listed below.

   - Some words of the input sentence do not match to words in target; all such words in the input sentence are valid dictionary words.
   - New words (valid dictionary words) are added or some words are missing in the input sentence.
   - Change in position of words leading to improper & grammatically inaccurate sentences.
   - Changing the sentence formation from direct to indirect speech or vice-versa; change of speech in sentences from first person to third person.

3. **Cat C:** The constituent words present in sentences contain spelling errors or typographic errors. The input sentences contain non-word errors and other tokens which are not found in dictionary. This category may contain words from languages other than English as well.

4. **Cat D:** This category or error consists of input sentences containing both Cat B & Cat C type errors. Thus, the sentences contain both non-word and real-word errors. For a language model, this is the most complex type of error to deal with due to the presence of both Cat B & C type errors. Such errors are very confusing to humans as well.

Table 1: Different Categories of Errors.

| Error Category | Description | Example Sentence | Example Target |
|---|---|---|---|
| Cat A | No errors detected in the text | These are cars. | These are cars. |

| Cat B | Grammatical errors detected in the text. | These are car. | These are cars. |
| Cat C | Spelling errors detected in the text. | These are rars. | These are cars. |
| Cat D | Sentence formation errors are introduced. | Thess are car. | These are cars. |

**Error Category Shift Causal Analysis**

There may be a mutlitude of reasons which causes a shift between any two categories of error. A systematic study has led to identification of these reasons which are tabulated in Table 2. It may be noted that these causes are identified based on study of the available dataset.

Table 2: List of causes leading to error category shift of sentences.

| | | Input | | | |
|---|---|---|---|---|---|
| | | Cat A | Cat B | Cat C | Cat D |
| Predicted | Cat A | • No Change | • Grammatical errors are corrected | • Spelling errors are corrected | • Both, spelling & grammatical errors are corrected |
| | Cat B | • New words introduced<br>• Words deleted<br>• Sequence change<br>• Existing correct word altered but no spelling error | • No change<br>• New words introduced<br>• Words deleted<br>• Sequence changes of existing words<br>• Existing correct word altered but no spelling error | • New words introduced<br>• Words deleted<br>• Sequence changes of existing words<br>• Existing correct word altered but no spelling error<br>• Spelling errors corrected to non-matching words | • New words introduced along with correction of erroneous words.<br>• Erroneous words deleted<br>• Sequence changes with spelling correction.<br>• Existing correct word altered but no spelling error<br>• Spelling errors corrected to non-matching words<br>• Spelling errors corrected |
| | Cat C | • New words with spelling errors<br>• Existing correct word altered but no grammatical error | • New words with spelling errors introduced with corrected grammar<br>• Sequence changed with correct grammar but spelling error | • No change.<br>• New spelling errors introduced.<br>• Existing spelling errors corrected with new spelling errors introduced. | • New words with spelling errors with corrected grammar<br>• Existing errors corrected but new words spelling errors<br>• Existing grammatical errors got corrected. |

|  |  | • Existing words corrected but misspelled |  | • Existing errors corrected but other existing words misspelled |
|---|---|---|---|---|
| Cat D | • New misspelt words and grammatical errors<br>• Existing words altered for spelling and grammatical errors<br>• Word deletion and existing correct words altered with spelling errors<br>• Sequence changed & existing correct words altered with spelling errors | • New misspelt words<br>• Some existing words misspelled<br>• Existing grammatical error corrected but new grammatical and spelling error introduced | • New words with grammatical and / or spelling errors<br>• Existing correct words sequence changed<br>• Existing spelling error corrected but new grammatical & spelling error introduced | • No change.<br>• New spelling OR / AND grammatical error added<br>• New misspelt words with corrected existing spelling error<br>• New grammatical error with corrected existing grammatical error |

## IV. Dataset

C4 Dataset is an Open-Source dataset obtained with Common Crawl web scrape [30]. This dataset contains millions of collected sentences along with their target sentences. The collected statements, referred to as input sentences in this work, belong to one of the categories defined in Sec. III. The input sentences are manually typed chat messages sent by users across the internet. These target sentences are manually inspected corrected input sentences by human experts. These target sentences act as the ground truth for this work. Due to constraints of available computing infrastructure, only a million sentences from this dataset are chosen for this work.

These one million sentences were analyzed for the error category distribution. Fig.1 represents the distribution pattern of the dataset chosen for this work. It may be observed that most of these input sentences are error free and belong to Cat A.

These one million sentences were split into train, validation, and test sets with 50:20:30 ratio. 500,000 sentences were used for training of the models while 200,000 sentences were used as validation sets (refer Table 3). A closer inspection of the test set sentences revealed that some of the target sentences contained errors (improper job in correcting the input sentences by human experts). Since the existence of such error affects the evaluation of the models, an exercise to inspect and correct the targets was undertaken. Such an exercise demands huge investment of time. Hence, a limited set of 25,738 sentences were inspected and corrected in this work. This set, called as 'UpdTest' is used for all testing and result reporting in this work.

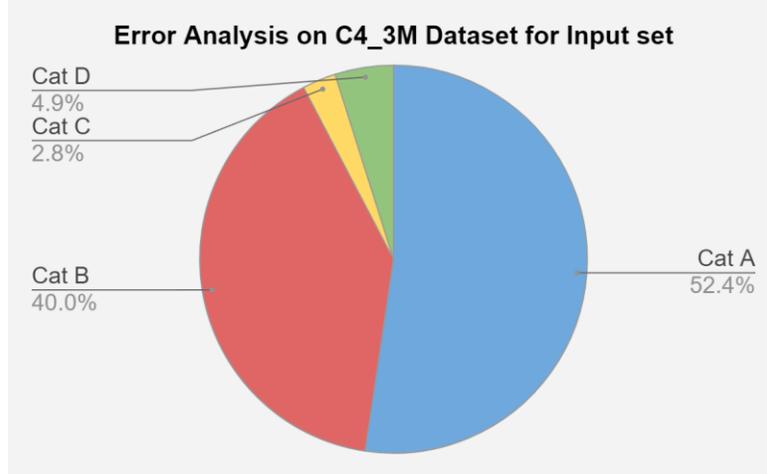

Fig. 1: Error category distribution of the input sentences in the dataset used for this work.

To understand the impact of target sentence inaccuracy on the model training, a study was conducted on the 25,738 samples in the UpdTest set. The study involved identification of sentences which passed as error free during the quality inspection process. It was observed that ~90% target sentences were error free. Since, most of the target sentences are error-free, it was assumed that only a negligible influence shall be exerted by the rest 10% of the erroneous sentences during the model training. So, a decision to continue with the training set (500,000 sentences) and validation set (200,000 sentences) was taken.

Table 3: Count of sentence records in Train, Test & Validation sets.

| Parameter | Value |
|---|---|
| Train | 500,000 |
| Validation | 200,000 |
| Test | 300,000 |
| UpdTest | 25,738 |

## V. Methodology
### a. Model Design

Sequence to Sequence (Seq2Seq) [31] models are an industry standard & used for various kinds of text analytics and natural language processing (NLP) tasks. Seq2Seq model employs advanced deep learning algorithms with elaborate encoder-decoder architectures to perform such tasks. The encoder in Seq2Seq models reads the input sequence and summarizes it into a fixed length representation called state or context vector. The decoder uses this context vector to generate an output which is task specific. The choice of architecture for each Seq2Seq model depends on the task at hand.

Seq2Seq models achieve their intended target tasks by conversion of textual data from one domain to another. Such conversion operations include machine translation [32], text summarization [33], image captioning [34], chatbot interactions [35] & query engines [36]. Generating error-free sentences from erroneous text is another such task which may be accomplished by Seq2Seq modeling [31]. This error correction is achieved by analysis of the context, neighborhood and meaning of the sentence. Two such Seq2Seq models, namely, Bidirectional Auto-Regressive Transformers (BART) and MarianMT, are utilized in this work.

BART is a self-supervised denoising autoencoder that aids in the recognition of words and phrases. It is a pre-trained model that combines the advantages of auto-regressive transformers with bidirectional traversal. BART's working consists of two sequential steps: (i) adding noise to the input text, & (ii) reconstructing the correct output. The first step is achieved either by corrupting a predetermined character or sequence, or by using appropriate noise-generating functions. The reconstruction of the correct output sequence is achieved with usage of a language model with a Seq2Seq architecture that aids in learning and reconstruction of the output string by substituting valid tokens for noise [37]. BART's architecture allows for selection of different nosing functions for effective learning. The BART model from the SimpleTransformers library [38] is utilized for this work.

MarianMT [39], a sophisticated neural translation framework that uses a sequence-to-sequence model, is mostly employed for text translation. They are C++ translational frameworks that are effective and independent. The foundation of MarianMT is also a typical encoder-decoder design. In the current work, MarianMT is used to transform incorrect English sentences into their corrected equivalents. It's assumed that such a transformation is equivalent to language translation task where incorrect sentences form a variation of the English language. The MarianMT model employed in this work uses pre-trained models from the SimpleTransformers library [38]. The decoder, "Helsinki-NLP/opus-mt-NORTH EU-NORTH EU," is trained with transfer learning in this work for converting sentences from erroneous English language back to corrected English, thereby, reducing textual error.

Both the pretrained models, BART and MarianMT, are trained on a train set of 500,000 records with a batch size of 32 sentences in each batch. The model employs beam search with a beam width of 5 for the prediction of tokens from sequence of words. A fivefold cross validation was also employed during the training process. The training lasted for ~27 hours on a GTX Titan Black GPU with 12 GB RAM to complete one epoch.

  b. *Error Category Analysis Algorithm*

To analyze the error category for the input sentences, an algorithm is developed. The sentences along with the constituent word tokens of the input-target pairs are utilized to accomplish the category determination task. A pseudo-code for the algorithm is presented in Fig. 2 while Fig. 3 provides a flowchart demonstration.

At the first stage of the algorithm, the input sentence is compared against the target to determine for their match. If the input matches the target sentence, then it belongs to Cat A. For non-match scenarios, both the input and target sentences are tokenized to extract their constituent words. The constituent words of the input sentence are searched for their existence in the target list as well as a dictionary. The input sentence word tokens which are not found in the target sentence are referred to as mis-matched words. If all the mis-matched words from the input are found to exist in the dictionary, then the input sentence is categorized as Cat B, considering it as a mere grammatical error. However, if some of these words are found to exist in the dictionary, then the sentence is declared as Cat D. If none of the mis-matched words are found in the dictionary, the mis-matched words in the input sentence are replaced with words corresponding to the index of the target sentence. If the target now matches the input, it is declared as Cat D else the input sentence is assumed to be containing pure typographic errors and declared as Cat C.

```
Note: Subscript represents index number (for example
Mismatched_words_(i,0) is same as Mismatched_words[i][0])

K ← 0
O ← 0
Mismatched_words ← [ ]

If Input matches Target
        Error category ← 'Cat A'
Else
        A ← Tokenized input
        B ← Tokenized target

For i ← 0 to (length of A)
    For j ← 0 to (length of B)
        If A_i is equal to B_j
            A_i=int(0)
            B_j=int(0)
            Break the inner loop
        Else
            K ← K+1

If K is equal to (length of B)
    Mismatched_words ← Mismatched_words + (A_i, i)

If length of mismatched_words is equal to 0
    Error category ← 'Cat B'
Else
    For i ← 0 to (length of Mismatched_words)
        If Mismatched_words_(i,0) belong to dictionary
            O ← O + 1
    If length of O is equal to length of Mismatched_words
        Error category ← 'Cat B'
    Else if length of O is less than length of Mismatched_words but not 0
        Error category ← 'Cat D'
    Else if length of O is equal to 0
        For H ← 0 to (length of Mismatched_words)
            U ← Mismatched_words_(h,1)
            Replace input_U with target_U
        If input matches target
            Error category ← 'Cat D'
        Else
            Error category ← 'Cat C'
```

Fig. 2: Pseudo-code for the Error Analysis

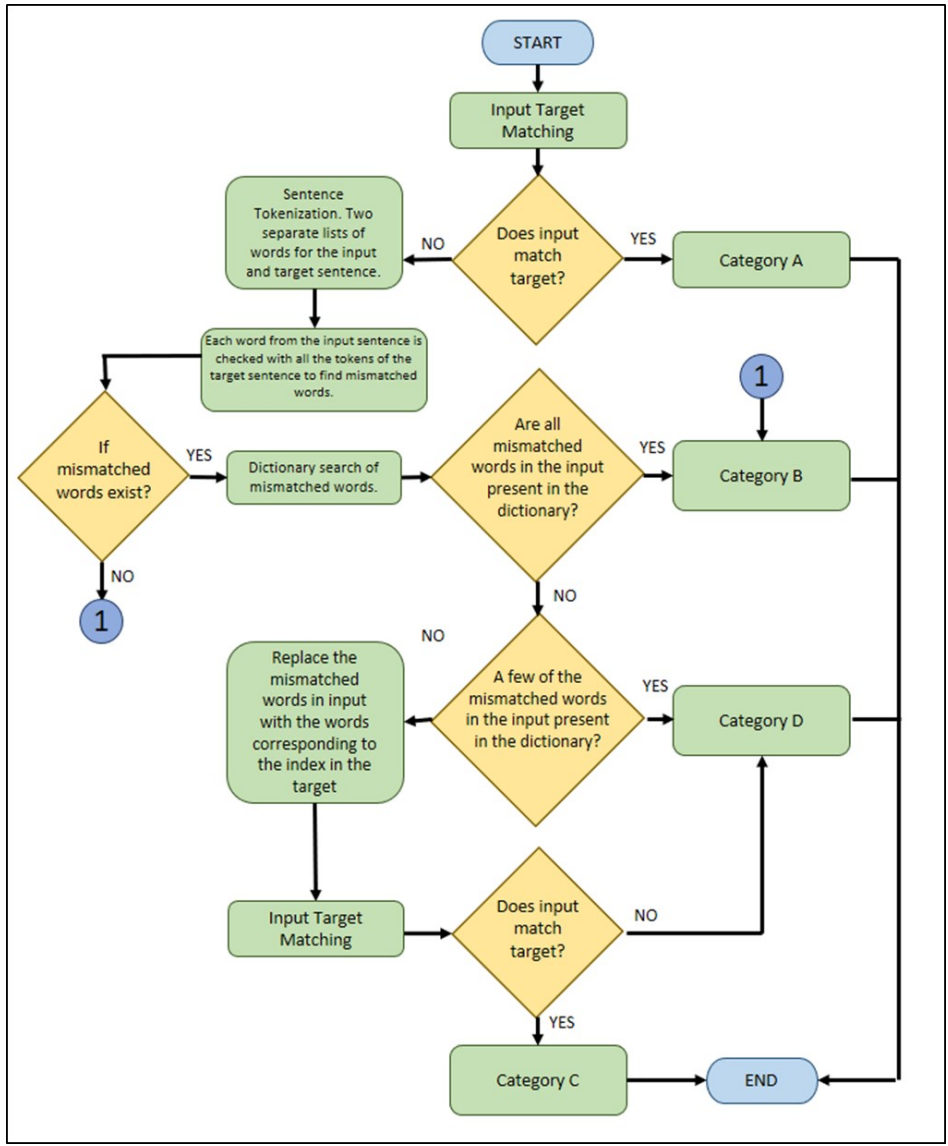

Fig. 3: Error analysis algorithm flow diagram

## VI. Result and discussion

Both the BART and MarianMT pre-trained models are trained with available trained data. Since the pretrained models are trained on billions of text samples, they already are capable of handling grammatical errors to varied degrees. Most of the prior works employ transfer learning for spelling error correction. Therefore, it is assumed that the model's ability to handle spelling errors is not adequate and transfer learning, with the training dataset, is performed to enhance these models' capacity in handling spelling errors.

*a. BART confusion matrix*

The predictions of the BART models for the test set (25,738 sentences) were analyzed with the error category detection module. The categories of error were recorded and tabulated as seen in Table 4. An analysis of error correction for different categories of input sentences by the BART model reveals that the model behaves quite differently while responding to each of them. While 99.8% of Cat A sentences were predicted correctly (sentences predicted as-is), only 9% of Cat B, 22% of Cat C and a miniscule of 3% of Cat D were fully corrected. Looking at this pattern, it may be inferred that BART is able to handle spelling error corrections better than grammatical errors.

Table 4: Confusion matrix for BART

|  |  | Input BART | | | | |
| --- | --- | --- | --- | --- | --- | --- |
|  |  | Cat A | Cat B | Cat C | Cat D | Total |
| Predicted | Cat A | 10975 | 1229 | 123 | 54 | 12381 |
|  | Cat B | 13 | 11169 | 66 | 449 | 11697 |
|  | Cat C | 0 | 2 | 324 | 24 | 350 |
|  | Cat D | 0 | 26 | 22 | 1262 | 1310 |
|  | Total | 10988 | 12426 | 535 | 1789 |  |

Similarly, for input sentences from Cat B, 89.9% of the sentences did not have any category shift. Only 1229 (9.9%) of sentences got corrected and moved to Cat A, while 28 (remaining 0.2%) sentences had spelling errors introduced, thereby pushing them to Cat C or D. This shift is due to the new spelling errors being induced. Thus, its inferred that the model introduces spelling errors, though no plausible explanation exists, currently, for such an introduction.

It may be observed from Table 4 that 123 (22.9%) sentences from Cat C were shifted to Cat A with complete elimination of spelling errors. The spelling errors of 66 (12.3%) sentences were corrected and these sentences were shifted to Cat B. This scenario happens when there are no spelling errors but the sentences contain grammatical errors. While category shift did not occur for 324 (60.5%) sentences (remained with Cat C as-is), predictions for 22 (4.1%) sentences moved them to Cat D. Various cases of movement from Cat C to Cat B or D are listed in Table 2.

With a 3% complete correction rate for Cat D, 54 sentences are predicted completely error free (a shift from Cat D to Cat A). Spelling errors are removed for 449 (25.2%) sentences (additionally, new grammatical errors might have been introduced), thereby shifting them to Cat B from Cat D. Furthermore, 24 (1.3%) sentences are moved from Cat D to Cat C. This occurs when either the grammatical mistakes are corrected or the new words responsible for grammatical correction are introduced with spelling errors. 1262 (70.5%) sentences still contained both grammatical and spelling errors remaining in Cat D. The various causes for such movements are listed down in Table 2.

Furthermore, Cat D to Cat B shift would mean an additional increase in the percentage of typographic errors being corrected. Similarly, Cat D to Cat C shift would mean an additional increase in the percentage of grammatical errors being corrected. Hence, total percentage of errors being corrected for each of these categories are listed in Table 5. It may be noted from this table that the BART model has performed significantly well for spelling error correction when compared to grammatical error correction. This is in contrast to the observations by Alikaniotis & Raheja [9] where they state that such models are suited to grammatical error correction (GEC).

Table 5: Error percentage for BART.

| Error | Formula | Calculation | Percentage |
|---|---|---|---|
| Total Spelling Error Correction Percentage | ((Cat D -> Cat B) + (Cat C -> Cat A))/ (Sum of Cat D transitions + Sum of Cat C transitions) | 572 / 2324 | 24.6% |
| Total Grammatical Error Correction Percentage | ((Cat D -> Cat C) + (Cat B -> Cat A)) / (Sum of Cat D transitions + Sum of Cat B transitions) | 1253 / 14215 | 8.8% |
| Total Mixed Error Correction Percentage | (Cat D -> Cat A) / (Sum of Cat D transitions) | 54 / 1789 | 3.1% |

b. *MarianMT confusion matrix*

Similar to BART, MarianMT models were also analyzed with UpdTest set sentences (25, 738) for error categorization, and the results are presented in Table 6. A total of 10956 (99.7%) sentences are predicted as-is error free & remain in Cat A. However, grammatical error was introduced in a tiny number of 32 sentences with 0.3% shift to Cat B. This could have occurred due to new words added by the model, deletion of words or sequence change of the sentence.

Table 6: Confusion matrix MarianMT

| | | Input MarianMT | | | | |
|---|---|---|---|---|---|---|
| | | Cat A | Cat B | Cat C | Cat D | Total |
| Predicted | Cat A | 10956 | 669 | 93 | 26 | 11744 |
| | Cat B | 32 | 11617 | 24 | 390 | 12063 |
| | Cat C | 0 | 6 | 363 | 18 | 387 |
| | Cat D | 0 | 134 | 55 | 1355 | 1544 |
| | Total | 10988 | 12426 | 535 | 1789 | |

While 5.4% of sentences with grammatical error were corrected and 669 sentences shifted from Cat B to Cat A, 11617 (93.5%) of sentences remained in Cat B as-is. A minute 0.04% (6 sentences) were shifted to Cat C with only spelling errors in them. 134 (1.1%) sentences with grammatical errors got additional spelling errors induced and shifted from Cat B to Cat D.

For input sentences in Cat C, 93 (17.4%) sentences got spelling errors rectified and shifted to Cat A while 363 (67.9%) sentences were predicted as-is & remained in Cat C. Spelling errors were corrected for 24 (4.5%) sentences to grammatical errors making a shift from Cat C to Cat B and 55 (10.3%) of the sentences shifted from Cat C to Cat D.

With a closer inspection, it may be noted from Table 6 that 26 (1.5%) sentences containing both grammatical and spelling errors were completely rectified and shifted to Cat A from Cat D. While 390 (21.8%) sentences were predicted spelling error free with shift to Cat B, 18 (1%) sentences were predicted grammatical error free and shifted to Cat C from Cat D. The reasons for such shifts are mentioned in Table 2. Finally, 1355 (75.7%) sentences remained as-is in Cat D.

Similar to BART, if we conduct an analysis on the Cat D to Cat B jump and Cat D to Cat C jump for MarianMT to assess the amount of grammatical and spelling errors being corrected, respectively. The results of the analysis are presented in Table 7. It may be noted from Table 7 that the total amount of corrections performed from MarianMT is less than from BART model. However, a similar pattern of correction with maximum corrections coming from spelling errors is observed for MarianMT as well.

Table 7: Error percentage for BART.

| Error | Formula | Calculation | Percentage |
|---|---|---|---|
| Total Spelling Error Correction Percentage | ((Cat D -> Cat B) + (Cat C -> Cat A))/ (Sum of Cat D transitions + Sum of Cat C transitions) | 483/2324 | 20.8% |
| Total Grammatical Error Correction Percentage | ((Cat D -> Cat C) + (Cat B -> Cat A)) / (Sum of Cat D transitions + Sum of Cat B transitions) | 687/14215 | 4.8% |
| Total Mixed Error Correction Percentage | (Cat D -> Cat A) / (Sum of Cat D transitions) | 26/1789 | 1.4% |

c. *Error Shift Analysis*

This subsection conducts a detailed analysis for all different categories of error shifts. While granular quantitative analysis for each cause for such shifts is not carried out, general pattern analysis, for multiple causes for such shifts, is performed here.

The errors introduced, change the structure and context of the sentence. The main idea behind analyzing the error and shift in error category is to study and identify the patterns associated with the shifts and understand the weak points of the models. This understanding may help, in future works, to increase the performance and accuracy of the models.

**Error Shifts for Cat A Input**

This category deals with input sentences which match perfectly to the target sentences. A shift from Cat A to any other category indicates that the predicted sentence contains error. Such change is possible due to one or more of the following causes: (i) errors, spelling or grammatical, exist in target while the predicted text is error-free, and (ii) models' predicted text is correct but an alternate form of sentence (output & target text mismatch but correct).

It may be observed in Table 4 & Table 6 that both BART and MarianMT have shifted sentences from Cat A to Cat B. Additionally, it may be recalled from the earlier discussion that most of the input sentences stayed put in the same category, thereby, confirming the assumption that both the chosen models are fairly resistant to inducing error for accurate input sentences.

*Category A -> Category B*

This subsection examines the cases of shifts from Cat A to Cat B with examples.

Table 8: Error category shift for Cat A to Cat B for BART and MarianMT.

|  | **Example 1** | **Example 2** | **Example 3** | **Example 4** |
|---|---|---|---|---|
| **target** | Because it **smells artistic**. | Enjoy this rather short **( 1' 40'' )** : It is the impressive story of a brave teenager rescuing dolphins. | The way of learning: reading **vs.** experience vs. experience of someone | I listened to an audio book `` **Doctor WHO** `` |
| **input** | Because it **smells artistic**. | Enjoy this rather short **( 1' 40'' )** : It is the impressive story of a brave teenager rescuing dolphins. | The way of learning : reading **vs.** experience vs. experience of someone | I listened to an audio book `` **Doctor WHO** `` |
| **BART** | Because it **smell** like **art**. | Enjoy this rather short **( 1'40'' )** : It is the impressive story of a brave teenager rescuing dolphins. | The way of learning : reading**,** experience vs. experience of someone | I listened to an audio book `` **Doctor WHO** `` |
| **Marian MT** | Because it **smell art**. | Enjoy this rather short **( 1'40'' )** : It is the impressive story of a brave teenager rescuing dolphins. | The way of learning : reading **vs.** experience vs. experience of someone | I listened to an audio book **Doctor WHO** |

Example 1 in Table 8 contains a sentence which starts with a conjunction, namely, "because". Since the input begin with a conjunction, it's extremely difficult to determine the context for the sentence. This poses a serious challenge in front of both BART and MarianMT to arrive at the intended meaning of the sentence, leading to failure of both these models. Example 2 presents a sentence which contains data in Degree, Minutes and Seconds (DMS) format. While definite symbolism for representation of such data exists, there is no standard framework which specifies usage of space between values for such data. Thus, data of this format exists both with and without spaces. Possibly, pre-trained models of both BART & MarianMT were trained on format of no-space representation for DMS. This led to both these models removing the space in the data. While the predicted sentences are not erroneous, yet they don't match the target, leading to shift in error category. A closer look at example 3 gives us the understanding that the BART model has introduced error by removing the word 'vs' whereas MarianMT's prediction has matched the target. BART predicted the sentence in an accurate, yet different representation. This can be attributed to the context understanding of the given input sentence by BART model. In Example 4, the input sentence contains a par of the sentence within 2 single quotes instead of a double quote. While this form of representation is not quite standard, its not completely erroneous either. BART has left the sentence untouched while MarianMT has removed the 2 single quotes. So, MarianMT generated a greater error in such prediction leading to a category shift.

To summarise, the change in error category is observed due the models' failure to understand the context of the sentence. Additionally, error-shifts are also observed for incomplete or improperly formed sentences.

**Category B Error Shifts:**

In the Category B error input text is erroneous as it consists of grammatical errors which change the structure and context of the sentence. These errors include omission or addition of words/phrases, change in parts of fundamental grammar. The following subsections examine, with examples, the cases of shifts for input statements in Cat B.

*Category B -> Category A*

In Example 1 of Table 9, "it" has been added by both the models to complete the sentences and make it error free. Similarly in example 2, the models added the article "the" to predict correct sentences. In example 3, BART corrects "I had been" to "I have been" to make the sentence contextually right. In example 4, in "most of people" "of" is omitted by MarianMT to predict a correct sentence.

Table 9: Error category shift for Cat B to Cat A for BART and MarianMT.

|  | **Example 1** | **Example 2** | **Example 3** | **Example 4** |
|---|---|---|---|---|
| **target_new** | I enjoyed **it.** | About 30 people joined **the** same company together. | So now **I have been** forcing myself to do it. | **Most people** think that Poland is a backward country. |
| **Input** | I enjoyed. | About 30 people joined same company together. | So now **I had been** forcing myself to do it. | **Most of people** think that Poland is a backward country. |
| **BART** | I enjoyed **it.** | About 30 people joined **the** same company together. | So now **I have been** forcing myself to do it. | **Most of people** think that Poland is a backward country. |
| **MarianMT** | I enjoyed **it.** | About 30 people joined **the** same company together. | So now **I had been** forcing myself to do it. | **Most people** think that Poland is a backward country. |

From the above examples, it can be observed that both the models understand the grammatical rules well enough to add articles wherever necessary. The models also interpret the correct tenses for simple input sentences. However, each model also have their failure scenarios though no specific pattern is observed for such failures.

*Category B -> Category C*

This subsection examines the cases of shifts from Cat B to Cat C with examples. In Table 10-Example 1, BART grammatically corrects "as a" to "as an" but misspells "orthodontist". Similarly in Example 2, "videos" is added for the subject verb agreement, but "imaginative" is misspelled as "imaginationful" by BART. MarianMT misspells "visit" as "visite" in Example 3. In Example 4, both the models changed "broadcast" to "broadcasted" and predicted sentences. However, its quite difficult to state if the models predicted wrongly given the constitution of the input sentence.

Table 10: Error category shift for Cat B to Cat C for BART and MarianMT.

|  | **Example 1** | **Example 2** | **Example 3** | **Example 4** |
|---|---|---|---|---|
| target_new | I think it's difficult for me to work **as an orthodontist.** | Their music **videos** are so COOL and they are very **imaginative.** | I would like to **visit** the USA. | It **was about** the Netherlands. |
| input | I think it's difficult for me to work **as a orthodontist.** | Their music **video** are so COOL and they are very **imagination**. | I would like to **visiting** USA. | It **broadcast about** the Netherlands. |
| BART | I think it's difficult for me to work **as an orthodist.** | Their music **videos** are so COOL and they are very **imaginationful**. | I would like to **visit** the USA. | It **was broadcasted** in the Netherlands. |
| MarianMT | I think it's difficult for me to work **as a orthodontist.** | Their music **video** are so COOL and they are very **imaginated.** | I would like to **visite** the USA. | It **broadcasted about** the Netherlands. |

The models are observed to have learnt well on grammatical rules like adding articles. But when they encounter possibly new words such as "orthodontist", spelling errors arise. This pattern of spelling error injection is also observed for small, incomplete and context-absent sentences.

*Category B -> Category D*

This subsection examines the shifts from Cat B to Cat D, with examples. Table 11 presents some example cases of this category shift. As may be observed in Example 1, "to a" was omitted in the input & the same was added by BART. Additionally, BART inserted a space between "Shijo" & "Karasuma" in the input word "ShijoKarasuma" which is a right thing to do. However, due to error in the target sentence, this is considered as an erroneous category shift. On the other hand, MarianMT not only misses to include "to a" into the sentence, it also incorrectly converts "ShijoKarasuma" to "Shijo". In Example 2, BART predicts "detail" as "TheDetail" as one word making it a spelling error and does not make "detail" plural. Except for the missing space, the sentence predicted by BART is correct, but a mismatch against the target. In Example 3, "Boeing" is misspelt as "teaching" and "byborg". The tense is also predicted wrong by MarianMT for the same sentence. Example 4 presents a case where both the models have failed to capture the grammar from the contest. Additionally, MarianMT predicts "IBT" as "VIT" which is change of the proper noun.

Table 11: Error category from Category B to Category D for BART and MarianMT

|        | **Example 1** | **Example 2** | **Example 3** | **Example 4** |
|--------|---------------|---------------|---------------|---------------|
| target | The day before yesterday, my old classmates and I went **to a** sushi bar at ShijoKarasuma. | **Details are in the** following site. | The Boeing 787, a new **aircraft** manufactured by **Boeing, came** to Japan today. | **Has** anyone **taken** the **IBT** toefl before? |
| input  | The day before yesterday, my old classmates and I went sushi bar at ShijoKarasuma. | **Detail is in** following site. | Boeing 787, a new **air craft** manufactured by **Boeing, come** to Japan today. | **Is** anyone **take** the **IBT** toefl before? |
| BART   | The day before yesterday, my old classmates and I went **to a** sushi bar at **Shijo Karasuma**.. | **TheDetail is in the** following site. | Boeing 787, a new **air craft** manufactured by **Boeing, came** to Japan today. | **Is** anyone **take** the **IBT** toefl before? |
| MarianMT | The day before yesterday, my old classmates and I went sushi bar at **Shijo** | in the following website. | **teaching** 787, a new **air craft** manufactured **byborg, come** to Japan today. | **Is** anyone **take** the **VIT** toefl before? |

In the above examples, the models predict correct sentences but do not match the target, or grammatically erroneous sentences due to lack of context. Additionally, these models add spelling errors with proper nouns, as they haven't learnt these words earlier.

**Category C Error Shifts:**

In the Category C error input text is erroneous as it consists of spelling errors which change the structure and context of the sentence. The spelling errors include words that cannot be found in the English dictionary, capitalization of proper nouns and abbreviation.

*Category C -> Category A*

Table 12: Error category from Category C to Category A for BART and MarianMT

|        | **Example 1** | **Example 2** | **Example 3** | **Example 4** |
|--------|---------------|---------------|---------------|---------------|
| target | Good **evening!** | I **want** to talk to her | So, I will study hard **tomorrow**. | Our daily life is getting **useful** day by day. |
| input  | Good **eveninng**! | I **wnat** to talk to her | So, I will study hard **tommorow**. | Our daily life is getting **usuful** day by day. |
| BART   | Good **evening!** | I **want** to talk to her | So, I will study hard **tomorrow**. | Our daily life is getting **usuful** day by day. |
| MarianMT | Good **evening!** | I **want** to talk to her | So, I will study hard **tommorow**. | Our daily life is getting **useful** day by day. |

This subsection examines the cases of shifts from Cat C to Cat A with examples. Table 12 presents some representative samples for such error category shift. In Example 1 and Example 2, 'evening' is misspelt as 'eveninng' and 'want' is misspelt as 'wnat', respectively. Both BART and MarianMT are able to handle these errors effectively and correct the sentences. In Example 3, BART corrects "tommorow" as "tomorrow" while MarianMT fails to achieve the same. Similarly in Example 4, MarianMT spell corrects "usuful" to "useful" and BART is not able to correct this spelling error.

Both BART and MarianMT models have been able to correct spelling errors in the above mentioned examples. This may be due the fact that the input sentences are simple, complete and with minimal spelling errors. Once again, there is no pattern to the observed failures.

*Category C -> Category B*

This subsection examines the cases of shifts from Cat C to Cat B with examples. Table 13 presents example cases of such shift. As shown in Example 1, BART corrected "it raind" to "it's been raining". This change makes the predicted sentence a mismatch with the target, thereby, declaring it as a grammatical error. MarianMT, on the other hand, changes "raind" to "rains" which makes an inaccurate as well as grammatically incorrect statement. In Example 2, both the models corrected "everydays" as "every day", treating it as two different words. Since both "everyday" and "every day" are valid forms of usage, the correction is appropriate, yet mismatches the target. In Example 3, BART puts "motivation" instead of "exercise" which not only alters the context of the sentence, but also marks it as grammatical error. Both models have failed here. Similarly in Example 4, MarianMT changes "sennd" to "send" instead of "second" which does not match the target while BART is able to correct the error.

Table 13: Error category from Category C to Category B for BART and MarianMT

|  | **Example 1** | **Example 2** | **Example 3** | **Example 4** |
|---|---|---|---|---|
| **target** | Today, it **rained** for a long time. | I use it for studying everyday. | I need more **exercise**. | The **second** one is the Chinese Pavilion in the EXPO. |
| **input** | Today, it **raind** for a long time. | I use it for studying **everydays.** | I need more **excersise**. | The **sennd** one is the Chinese Pavilion in the EXPO. |
| **BART** | Today, **it's been raining** for a long time. | I use it for studying **every day.** | I need more **motivation**. | The **second** one is the Chinese Pavilion in the EXPO. |
| **MarianMT** | Today, it **rains** for a long time. | I use it for studying **every day.** | I need more **excersise**. | The **send** one is the Chinese Pavilion in the EXPO. |

Both the models are fairly able to capture and correct spelling errors in sentences with adequate context. Sometimes in the process of correction, especially for shorter sentences lacking contextual information, the corrections lead to formation of either alternate forms of sentences or different sentences.

*Category C -> Category D*

This subsection examines the cases of shifts from Cat C to Cat D with examples. Table 14 presents selected examples from such category shift.

Table 14: Error category from Category C to Category D for BART and MarianMT

|  | **Example 1** | **Example 2** | **Example 3** | **Example 4** |
|---|---|---|---|---|
| **target_new** | I sometimes felt sea **sickness.** | Fortunately **everyone was** on time! | it's even **offensive** in Japan. | But I'll try to keep writing journals regularly from now on. |
| **input** | I sometimes felt sea **sichness.** | Fortunately **eveyone cames** on time! | it's even **offence** in Japan. | But I'll try to keep **writng jounal** regularly from now on. |
| **BART** | I sometimes felt **the** sea **sichness.** | Fortunately **eveyone came** on time! | it's even **an offence** in Japan. | But I'll try to keep writing **jounal** regularly from now on. |
| **MarianMT** | I sometimes felt **seakelness**. | Fortunately **eveyone came** on time! | it's even **offence** in Japan. | But I'll try to keep writing **a jounal** regularly from now on. |

As shown in Example 1, BART failed to correct "sichness" to "sickness". It also added an unwanted article "the". Similarly in Example 3, BART added the article "an" making the sentence grammatically correct but a mis-match against the target. In Example 2, both the models corrected "cames" to "came" making the sentence correct. A closer examination suggests the target to be inaccurate representation for the input sentence. In Example 4, both BART & MarianMT corrects the spelling error in "writng" to "writing" but fails to handle the same for "jounal".

Though the models have failed to correct the spelling errors in this scenario, they may have generated better versions of the sentences in a grammatical sense. This could be due to lack of context or incomplete sentences given as input.

**Category D Error Shifts:**

In the Category D error input text is erroneous as it consists of both spelling errors and grammatical errors which change the structure and context of the sentence.

*Category D -> Category A*

This subsection examines the cases of shifts from Cat D to Cat A with examples. As seen in Example 1 of Table 15, both the models correct the spelling of "people" and change "problem" to its plural form to generate an error free sentence. Similarly, in Example 2, "of course" is corrected by both the models. In example 3, BART corrects the spelling of "visit" and adds "the" to give a completely correct sentence. MarianMT on the other hand has been able to add the article "the" but fails to correct the spelling error. In Example 4, MarianMT has been able to correct the spelling error in "communicating" as well as convert the "foreigner" to its plural form. BART, here on the other hand, has failed on both the accounts.

The models have learnt well to adjust the words in singular or plural according to the context as well as to add required articles like "the" wherever required. Additionally, the models also made spelling corrections.

Table 15: Error category from Category D to Category A for BART and MarianMT

|          | **Example 1** | **Example 2** | **Example 3** | **Example 4** |
|----------|---------------|---------------|---------------|---------------|
| **target** | So younger **people** in Japan should have more interest in these **problems**. | **Of course** I should speak in English there! | I will **visit the** USA this winter! | The Internet helps us with **communicating** with **foreigners**. |
| **input** | So younger **peopole** in Japan should have more interest in these **problem**. | **Ofcourse** I should speak in English there! | I will **visite** USA in this winter! | The Internet helps us with **communicatig** with **foreigner.** |
| **BART** | So younger **people** in Japan should have more interest in these **problems**. | **Of course** I should speak in English there! | I will **visit the** USA this winter! | The Internet helps us with **communicatig** with **foreigner.** |
| **MarianMT** | So younger **people** in Japan should have more interest in these **problems**. | **Of course** I should speak in English there! | I will **visite the** USA in this winter! | The Internet helps us with **communicating** with **foreigners.** |

*Category D -> Category B*

This subsection examines the cases of shifts from Cat D to Cat B with examples. Table 16 presents some chosen examples to illustrate the patterns of shifts observed in this category.

Table 16: Error category from Category D to Category B for BART and MarianMT

|          | **Example 1** | **Example 2** | **Example 3** | **Example 4** |
|----------|---------------|---------------|---------------|---------------|
| **target** | He **meets** her **everyday.** | So I can speak **a little** French. | Luckily, **I've found** this **website.** | Today my **sister came** to see me. |
| **input** | He **meet** her **evryday**. | So I can speak French a **littele.** | Luckly, I **find** this **web.** | Today my **sisiter come** to see me. |
| **BART** | He **met** her **yesterday.** | So I can speak French a **little.** | Luckily, **I** found this website. | Today my **sisiter came** to see me. |
| **MarianMT** | He **meet** her **everyday.** | So I can speak French a **little.** | Luckily, **I find** this website. | Today my **sister come** to see me. |

As seen in Example 1, MarianMT has been able to correct the spelling error but fails to correct the grammatical error. BART, on the other hand, predicts a perfectly correct but different statement with interpretation of "evryday" as "yesterday". In Example 3, BART predicted the sentence is contextually right, but a mismatch to the target. MarianMT corrects "website" but the sentence presents a meaning which is different from the intended one. In Example 2, both the models correct little's spelling, but produce sentences with grammatical error. In example 4, MarianMT corrects sister's spelling but the tense used is wrong.

The above examples demonstrate that the models predict sentences that remain the same, that are contextually correct but do not match the target or the model fails to learn the context and produces a different and wrong sentence. It's observed that BART performs superior corrections when lengthier sentences with better context is presented.

*Category D -> Category C*

This subsection examines the cases of shifts from Cat D to Cat C with examples. Table 17 presents example cases to illustrate patterns for this category shift.

Table 17: Error category from Category D to Category C for BART and MarianMT

|            | **Example 1**                  | **Example 2**                              | **Example 3**                                 | **Example 4**                  |
|------------|--------------------------------|--------------------------------------------|-----------------------------------------------|--------------------------------|
| target_new | I was **very surprised.**      | If you learn **Japanese, I will** help you.| Japanese **believe** that it is a **lucky** symbol. | It will be **a nice** time.    |
| input      | I was **suprised very** much.  | If you learn **Japanease, I help** you.    | Japanese **believes** that it is a **luckey** symbol. | It will be **naice** time.     |
| BART       | I was **very suprised.**       | If you learn **Japanase, I will** help you.| Japanese **believe** that it is a **luckey** symbol. | It will be **naice** time.     |
| MarianMT   | I was **very suprised**.       | If you learn **Japanese**, **I help** you. | Japanese **believes** that it is a **luckey** symbol. | It will be **a naice** time**.** |

As seen in Example 1 "surprised" spelling was not corrected by both the models. In Example 2, BART is unable to make corrections to spelling error present in the proper noun "Japanese" while MarianMT handles this properly. On the other hand, MarianMT is not efficiently able to handle the error in grammar while BART is able to correct the grammar error. In example 3, BART corrects "believe" to "believes" but does correct "luckey". In example 4, MarianMT makes the grammatical correction by adding "a", but "nice" spelling remains erroneous.

These models fail to rectify spelling errors because the words may be proper nouns or the model encounters these words for the first time, hence fails to understand its context and usage.

## VII. Conclusion

It can be concluded that both BART and MarianMT trained better for spelling error correction than grammatical error correction, with BART outperforming MarianMT. Both the models established some new errors while rectifying existing errors hence moving from grammatical error category to spelling error category and vice versa, it can be inferred that although these models correct the error present in the sentence there is a possibility that a new type of error be introduced. During the error category jump analysis, interesting scenarios were observed. Though the models did not predict the exact target sentence, they generated English sentences that retained the context of the statement. The most significant achievement of this analysis is the shift from Cat D to Cat A. While this shift happened only on 3% of the data, it can be concluded that BART and MarianMT has the ability to correct a mixed category of errors making it a robust model. A crucial observation was that the shift from Cat D to Cat B and Cat D to Cat C are more common, while the shift to Cat A is rare.

While understanding the analysis and conclusions drawn in this work, it can be extended further by conducting a detailed study of shifts in categories and comprehending the reason behind the shift in each sentence. The grammatical error category can be further split into multiple categories such as words omission, words addition, sequence change, grammar change, and an analysis can be performed to study the performance of the models better. A test to check the breaking point of the model can be conducted, to realize the limit of the model's ability to correct errors. Finally, this work can be expanded further to test with other language models to learn their strong points.

## DECLARATIONS

### Acknowledgement

The authors thank their university for the needed infrastructure support for this research work & manuscript preparation.

### Ethical Approval

Not Applicable.

### Availability of supporting data

The C4 dataset, used for the work reported in this paper, is available at https://www.tensorflow.org/datasets/catalog/c4.

### Competing interests

The authors declare that they have no competing interest for this work.

### Funding

No external funding was received for this work.